\documentclass[runningheads]{llncs}
\usepackage{graphicx}
\usepackage{mathrsfs}
\usepackage{amsmath,amscd,amsbsy,amssymb}
\usepackage{hyperref}

\DeclareMathOperator*{\argmin}{arg\,min}

\begin{document}
\title{A Neural Network Implementation for Free Energy Principle\thanks{This work is done in Prof. Shlomo Dubnov's experimental seminar.}}

%
\author{Jingwei Liu\inst{1}\orcidID{0009-0004-3160-9326}}
%
\authorrunning{J. Liu}
%
\institute{University of California San Diego, La Jolla CA 92092, USA}
\maketitle              
\begin{abstract}
The free energy principle (FEP), as an encompassing framework and a unified brain theory, has been widely applied to account for various problems in fields such as cognitive science, neuroscience, social interaction, and hermeneutics. As a computational model deeply rooted in math and statistics, FEP posits an optimization problem based on variational Bayes, which is solved either by dynamic programming or expectation maximization in practice. However, there seems to be a bottleneck in extending the FEP to machine learning and implementing such models with neural networks. This paper gives a preliminary attempt at bridging FEP and machine learning, via a classical neural network model, the Helmholtz machine. As a variational machine learning model, the Helmholtz machine is optimized by minimizing its free energy, the same objective as FEP. Although the Helmholtz machine is not temporal, it gives an ideal parallel to the vanilla FEP and the hierarchical model of the brain, under which the active inference and predictive coding could be formulated coherently. Besides a detailed theoretical discussion, the paper also presents a preliminary experiment to validate the hypothesis. By fine-tuning the trained neural network through active inference, the model performance is promoted to accuracy above 99\%. In the meantime, the data distribution is continuously deformed to a salience that conforms to the model representation, as a result of active sampling.

\end{abstract}
\section{Introduction}

Free Energy Principle (FEP) as an encompassing framework and a unified brain theory \cite{friston2010free} has been widely applied to account for various phenomena in many cognition, humanity-related fields such as psychology\cite{carhart2010default}, music\cite{koelsch2019predictive}, linguistic communication\cite{friston2020generative}, cultural niche construction\cite{constant2018variational}, embodiment\cite{allen2018cognitivism}, autopoiesis\cite{kirchhoff2018markov}, emotion recognition\cite{demekas2020investigation}. In the meanwhile, as a computational model deeply rooted in math and statistics, FEP posits an optimization problem based on variational Bayesian inference, which is solved either by dynamic programming\cite{friston2018deep} or expectation maximization\cite{friston2008hierarchical}. However, there seems to be a bottleneck in extending FEP to the fields of machine learning, which have been the hotspots for solving statistical and engineering problems in recent years. There are a few works that bridge active inference and reinforcement learning\cite{tschantz2020reinforcement}, and here is a survey on seeking the common ground for active inference and deep learning\cite{mazzaglia2022free}. However, as a variational-based method, FEP cannot be seen equally, or be generalized trivially, to reinforcement learning, a reward-based training method. This work gives a preliminary attempt at bridging FEP and machine learning, via a classical neural network model, the Helmholtz machine. There are three features of using the Helmholtz machine to study FEP under neural network settings:

\begin{enumerate}
\item The Helmholtz machine uses variational free energy as its objective, which is in accord with the FEP. In other words, we maintain the variational essence of FEP by using a corresponding variational machine-learning method, which minimizes the free energy.
\item If reinforcement learning is considered as capturing the qualities of expected free energy, which involves planning and future outcomes of sequential events, then the Helmholtz machine is a perfect parallel for the free energy. Although the Helmholtz machine is not temporal, in many aspects, it's an ideal prototype for implementing FEP and active inference in a neural network fashion, which will be argued extensively in this paper.
\item The Helmholtz machine also presents a satisfactory simulation for the hierarchical model of the brain. The forward and backward connections and hierarchical message passing are inherent in the implementation of the Helmholtz machine.
\end{enumerate}

This paper includes two main sections. Section 2 gives a theoretical account to the interrelationship of the free energy principle and the Helmholtz machine from aspects of mathematical formulation, model training and parameter updating, biological interpretability and plausibility. It provides a theoretical basis for generalizing FEP via the Helmholtz machine to broader model-fitting schemas in the neural networks for machine learning. Section 3 presents a preliminary experiment we designed to test the model. The model performs pretty well as the theoretical analysis indicates. In training stage I, the Helmholtz machine achieves an accuracy of 0.94 under traditional data fitting; In training stage II, we apply the active inference in FEP to actively sample the input sensations as salience. After a few rounds of fine-tuning, the model accuracy was boosted above 0.99, which presents high generation accuracy while keeping generation diversity at a satisfactory level.

\section{Free Energy Principle and Helmholtz Machine}

\subsection{Variational Inference for Statistics}

Here we give a brief overview of variational inference (VI)\cite{2016arXiv160100670B} in Bayesian statistics and show how this concept is linked to the free energy principle and Helmholtz machine. To maintain the notational consistency, we use the standard notations in \cite{dayan1995helmholtz}.

The goal is to determine the posterior $P(\alpha|d)$, where $d$ denotes the observable data, and $\alpha$ is variously referred to as hidden causes, latent variables, or hidden states. According to Bayes' rule,
\begin{equation}
    P(\alpha|d) = \frac{P(\alpha,d)}{P(d)} = \frac{P(\alpha,d)}{\int_\alpha P(\alpha,d) d\alpha}
\end{equation}
The integral over underlying causes is usually intractable (either unavailable in closed form or requires exponential time to compute), so the true posterior $P(\alpha|d)$ cannot be computed directly.

\begin{remark}
   We give a separate account for the conditional data $d$ in $P(\alpha|d)$. In classical variational inference\cite{2016arXiv160100670B}, $d$ represents the entire dataset, thus the latent distribution $P(\alpha|d)$ is independent of single data point, and all model properties resort to latent local and global variables. Therefore, we can see that an approximate posterior $Q(\alpha)$ is used to approximate the true posterior $P(\alpha|d)$, where $Q$ is not conditioned on the observations. This formulation is widely used in statistical variational inference and all resources I've read about the free energy principle such as \cite{FRISTON2016862} \cite{doi:10.1080/17588928.2015.1020053} \cite{parr2019generalised}.

   However, in many other settings, we frequently see that another form of approximate posterior $Q(\alpha|d)$ is used, the most prominent case is in VAE \cite{2013arXiv1312.6114K}. In the Helmholtz machine, this conditional $Q(\alpha|d)$ is also used as approximate posterior (it's not explicitly given in \cite{dayan1995helmholtz}, but in \cite{hinton1995wake}, it's clearly stated). The main reason is that the data $d$ is treated point-wisely in these models, thus the latent cause distributions conditioned on individual data points vary from each other. However, as the distribution $Q(\alpha|d)$ is parameterized by $\phi$, which is amortized to all data points, the approximate posterior is still tractable and works a similar way as in the unconditioned case.
\end{remark}

As the Helmholtz machine is the model we adopt, we will use the conditioned approximate posterior in the following discussions. In spirit, it differs little from the vanilla version as the formula deduction unfolds. To recap, we use an approximate posterior $Q_\phi(\alpha|d)$ to approximate the true posterior $P(\alpha|d)$, where $Q_\phi(\alpha|d)$ belongs to a parameterized family $\mathscr{Q}_\phi$ of probability densities. Our goal is to find the member of this family that minimizes Kullback-Leibler (KL) divergence to the exact posterior,
\begin{equation}
    Q_\phi^*(\alpha|d) = \argmin_{Q_\phi(\alpha|d) \in \mathscr{Q_\phi}} D_{KL}[Q_\phi(\alpha|d) || P(\alpha|d)]
\end{equation}

The variational method kicks in when we decompose the true posterior in the KL-divergence term,
\begin{align}
D_{KL}[Q_\phi(\alpha|d) || P(\alpha|d)] &= \mathbb{E}_Q[\log Q_\phi(\alpha|d)] - \mathbb{E}_Q[\log P(\alpha|d)] \\
                                        &= \mathbb{E}_Q[\log Q_\phi(\alpha|d)] - \mathbb{E}_Q[\log P(\alpha,d)] + \log P(d) \label{2}
\end{align}
The distribution $P(\alpha,d)$ is called the generative model, as it denotes the joint distribution of the latent and observable variables. The generative model is usually assumed as known in VI and FEP, as the way environment generates observations from causes is innate.

Now the problem falls on the third term in Equation (\ref{2}), what we call the log-evidence, or negative surprisal, $\log P(d)$. This term is again intractable, so to circumvent it, we use the nonnegativity of KL-divergence, rewrite (\ref{2}) as
\begin{equation}
    \log P(d) \ge \mathbb{E}_Q[\log P(\alpha,d)] - \mathbb{E}_Q[\log Q_\phi(\alpha|d)]
\end{equation}
where the right-hand side term $\mathbb{E}_Q[\log P(\alpha,d)] - \mathbb{E}_Q[\log Q_\phi(\alpha|d)]$ is called the evidence lower bound (ELBO). By maximizing ELBO we implicitly maximize the log-evidence $\log P(d)$. The free energy is given by the negative ELBO,
\begin{equation}
F = \mathbb{E}_Q[\log Q_\phi(\alpha|d)] - \mathbb{E}_Q[\log P(\alpha,d)] = D_{KL}[Q_\phi(\alpha|d) || P(\alpha,d)]
\end{equation}
which is the ultimate minimization goal in VI, FEP, and the Helmholtz machine.

\begin{remark}
The minimization term $F$ is seen as a compromise in VI. Since we cannot minimize $D_{KL}[Q_\phi(\alpha|d) || P(\alpha|d)]$ directly, we find some cheap approximation that we can compute. However, I claim it's not the case for generative models. In generative models which use generation as an organic component of model construction, the generative density $P(\alpha, d)$ is a necessity instead of the posterior $P(\alpha|d)$, since we are not only finding the best set of parameters in a density family that approximates a given distribution, but also using generated samples to regulate the recognition process (Helmholtz machine, VAE) or active sampling the generations to improve accuracy (FEP).
\end{remark}

In Helmholtz machine, instead of pre-defining a generative model $P(\alpha, d)$, in a more realistic way (since the generative density is usually unknown in real-life problems), we parameterize this distribution by $\theta$, and construct the free energy minimization goal
\begin{equation}
F = D_{KL}[Q_\phi(\alpha|d) || P_\theta(\alpha,d)] \label{helm}
\end{equation}
By jointly optimizing the two sets of parameters $\phi$ and $\theta$ in an EM (expectation-maximization) manner, we minimize the free energy of the system.

In FEP, the free energy is reformulated as the two equations,
\begin{align}
F &=  D_{KL}[Q_\phi(\alpha|d) || P(\alpha|d)] - \log P(d)  \label{3} \\
  &=  D_{KL}[Q_\phi(\alpha|d) || P(\alpha)] - \mathbb{E}_Q[\log P(d|\alpha)]\label{4}
\end{align}
Equation (\ref{3}) is interpreted from its first term as optimizing the recognition density of the brain to approximate the true distribution of the world, which shares the same goal as VI, minimizing $D_{KL}[Q_\phi(\alpha|d) || P(\alpha|d)]$; Equation (\ref{4}) is interpreted more inclined to its second term, $\mathbb{E}_Q[\log P(d|\alpha)]$, as a way to actively sample the sensory inputs that conform to the current representations, thus improving accuracy (please refer to \cite{friston2010free} for more details).                                            
In this work, we will integrate the classical Helmholtz machine which is trained under minimization of variational free energy with the active inference in FEP. Besides the parameter optimization, the model also performs active inference by a selective sampling of the environment, which entails a modulation of attention reflected in the distribution of evidence.

\subsection{Neural Network for Machine learning}
This work explores a way of implementing FEP using neural networks in machine learning. Traditionally, problems formulated under FEP are either solved by DEM (dynamic expectation-maximization)\cite{friston2009predictive} or MDP (Markov decision process)\cite{parr2019generalised}. Although the Helmholtz machine is not a temporal model, it uses real neurons specified by modern neural networks and updates its parameters via gradient descent. Under the current world trend, we believe it's imperative to extend FEP to the machine learning field and to solve problems using neural network architectures.

\begin{figure}
\centering
\includegraphics[scale = 0.65]{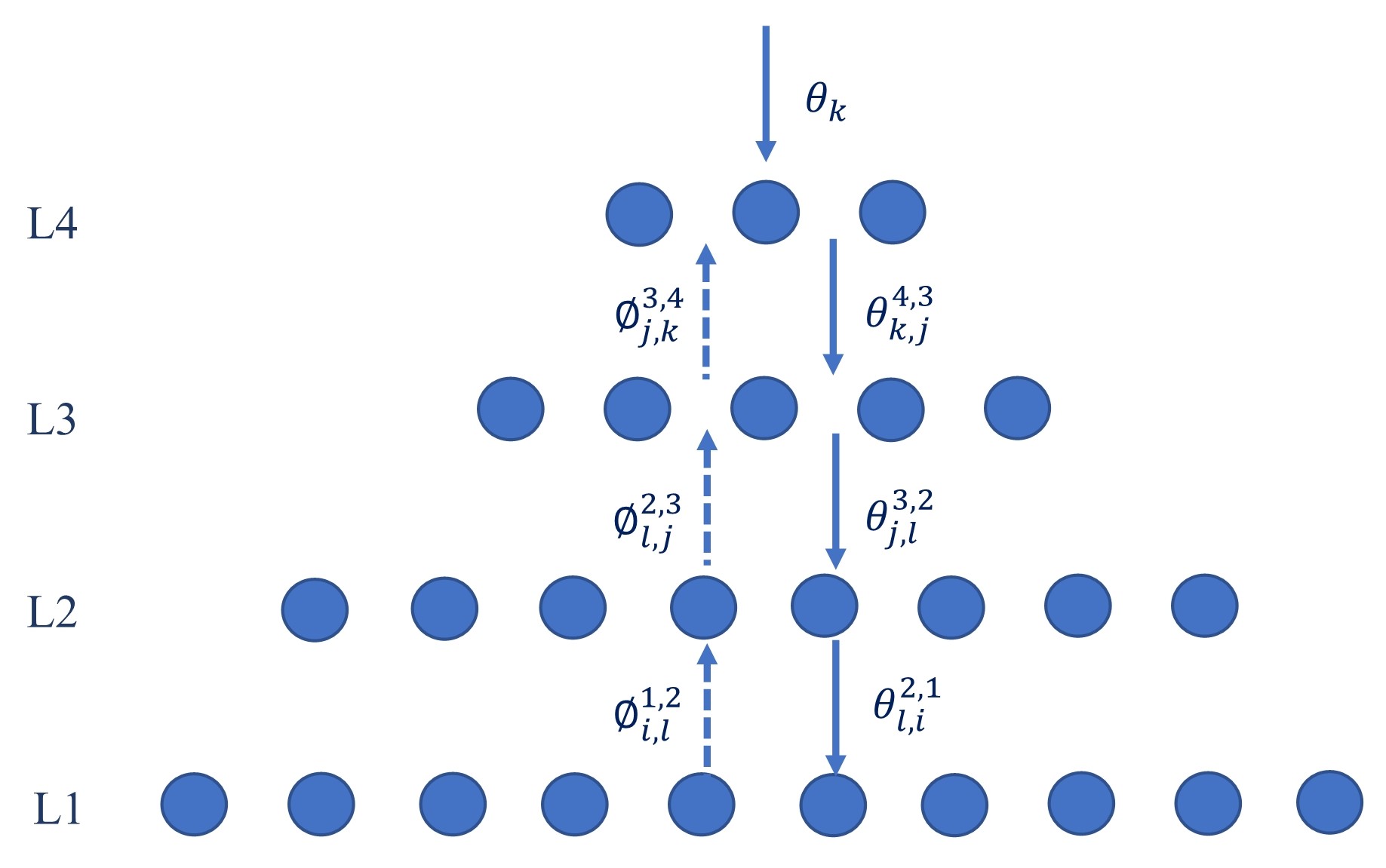}
\caption{The Helmholtz Machine. The Helmholtz Machine is a fully connected feedback neural network with hierarchical architecture. The solid lines show the bottom-up recognition process parameterized by $\phi$, and the dashed lines show the top-down generation process parameterized by $\theta$. The activity of each neuron is computed from the activities of all neurons in its previous layer. The activation functions are given in the text.} \label{fig1}
\end{figure}

The structure of the Helmholtz machine is shown in Fig. \ref{fig1}. It's a layered hierarchical model composed of stochastic binary neurons, connected by bottom-up recognition weights $\phi$ and top-down generative weights $\theta$. In this work, we did two major modifications to the original model in \cite{dayan1995helmholtz},

\begin{enumerate}
    \item The activity of the stochastic binary neuron is changed from $\{0,1\}$ to $\{-1,1\}$.
    \item The bias is added when computing the linear activation of each neuron.
\end{enumerate}

The first modification is done due to the derivative form of $F$ with respect to its parameters, for example, $\frac{\partial F}{\partial \theta_{k,n}^{m+1,m}} = -s_k^{m+1}(s_n^m-p_n^m)$. The neuron activity from the previous layer is used as a multiplier on the derivatives, which means if $s_k^{m+1} = 0$, the gradient equals zero, thus no updating will be performed for this parameter $\theta_{k,n}^{m+1,m}$ by gradient descent. Therefore, the parameter updating will be half paralyzed when neuron activities alternate between $0$ and $1$. To make the learning more efficient, we replace the activity value $0$ with $-1$, thus zero gradients won't occur unless $p_n^m$ approaches $s_n^m$.

The Helmholtz machine is fully connected. The activation of each neuron is a linear combination of all the neurons from its previous layer plus bias,
\begin{equation}
    a_n^m(\theta, \mathbf{s}^{m+1}) = \sum_k \theta_{k,n}^{m+1,m} s_k^{m+1} + b_{n}^{m+1,m}
\end{equation}
where the activation of the $n$-th neuron in layer $m$ is computed by a weighted sum of all activities $s_k^{m+1}$ in layer $m+1$ weighted by its corresponding parameter $\theta_{k,n}^{m+1,m}$, plus the bias $b_{n}^{m+1,m}$ for this neuron. Here the previous layer is the upper layer $m+1$, which corresponds to the top-down generative process indicated by dashed lines in Fig. \ref{fig1}. We added the bias term to the original formulation to expand the parameter set, thus endowing more freedom for the neural network to fit the data.

The probability is calculated by the sigmoid function $\sigma(x) = 1 / (1+e^{-x})$ of activation, which nonlinearly compresses the range into $(0,1)$.
\begin{equation}
    p_n^m(\theta, \mathbf{s}^{m+1}) = \sigma(a_n^m) = \sigma(\sum_k \theta_{k,n}^{m+1,m} s_k^{m+1} + b_{n}^{m+1,m}) \label{p}
\end{equation}
Similarly, the activation probability of a neuron in the bottom-up recognition process is computed as
\begin{equation}
    q_n^m(\phi, \mathbf{s}^{m-1}) = \sigma(\sum_k \phi_{k,n}^{m-1,m} s_k^{m-1} + b_{n}^{m-1,m}) \label{q}
\end{equation}
where the previous layer is the lower layer $m-1$, and the probability is denoted with $q$. The notations are consistent with our notations in Equation (\ref{helm}). The recognition density $Q_\phi(\alpha|d)$ and the generative density $P_\theta(\alpha,d)$ are computed by the product of the probabilities of all neurons, namely
\begin{align}
Q_\phi(\alpha|d) &= \prod_{m>1} \prod_n [q_n^m(\phi, \mathbf{s}^{m-1})]^{\frac{1+s_n^m}{2}} [1-q_n^m(\phi, \mathbf{s}^{m-1})]^{\frac{1-s_n^m}{2}} \label{Q}\\
P_\theta(\alpha,d) &= \prod_{m \ge 1} \prod_n [p_n^m(\theta, \mathbf{s}^{m+1})]^{\frac{1+s_n^m}{2}} [1-p_n^m(\theta, \mathbf{s}^{m+1})]^{\frac{1-s_n^m}{2}} \label{P}
\end{align}
Each neuron gives a Bernoulli distribution as the activity of $s_n^m$ takes value $-1$ or $1$.

As the explicit form of the free energy $F =  D_{KL}[Q_\phi(\alpha|d) || P_\theta(\alpha,d)]$ is given by equations (\ref{Q}) (\ref{P}), now we should consider how to compute the derivatives of $F_{\phi,\theta}$ thus minimizing this term by gradient descent. If we refer back to the structure of the Helmholtz machine in Fig. \ref{fig1}, we can see the neurons are connected recurrently. Besides that, the change of activities in one layer will affect the behavior of all neurons in higher layers, which makes backpropagation extremely difficult. To tackle this problem, a customized algorithm is designed for training this system, which is the wake-sleep algorithm \cite{hinton1995wake}. 

The wake-sleep algorithm disentangles the recognition parameters $\phi$ and the generative parameter $\theta$ by separating the training into two phases. In the wake phase, the bottom-up recognition process is performed using the current weights $\phi$ to get an instance of complete neuron assignments $\alpha$ by sampling activities from $\{-1,1\}$ based on the probability of each neuron $q_n^m$. Now we update the generative parameters $\theta$ based on the complete neuron activities encoded in $\alpha$, which makes the target values available for training the hidden units locally, thus disentangling them from the multi-layer coupling. In the sleep phase, the recognition weights $\phi$ are turned off and a random instance is generated based on the current top-down weights $\theta$. Alternatively, we update $\phi$ according to the neuron assignments of this generated instance while keeping the generative weights fixed. By iterating the wake and sleep phases, updating $\phi$ and $\theta$ alternatively, the objective function $F_{\phi,\theta}$ is minimized in an EM manner.

Now let's compute the derivatives of $F_{\phi,\theta}$ with respect to the generative weights. We can write out Equation (\ref{helm}) as
\begin{equation}
    F = \mathbb{E}_{Q_\phi}[\log Q_\phi(\alpha|d)] - \mathbb{E}_{Q_\phi}[\log P_\theta(\alpha,d)] \label{11}
\end{equation}
Since $\phi$ and $\theta$ are decoupled, the first term in Equation (\ref{11}) is constant when computing $\frac{\partial F}{\partial \theta}$, thus we write out the second term as
\begin{equation}
    \mathbb{E}_{Q_\phi}[\log P_\theta(\alpha,d)] = \sum_\alpha Q_\phi(\alpha|d)\log P_\theta(\alpha,d) \label{12}
\end{equation}
As the latent cause $\alpha$ is generated by random sampling as a single instance, the summation over all possible hidden causes in Equation (\ref{12}) is ignored with the weighting term $Q_\phi(\alpha|d)$, thus the objective is further simplified as 
\begin{equation}
    \log P_\theta(\alpha,d) =  \sum_{m \ge 1} \sum_n ({\frac{1+s_n^m}{2}} \log[p_n^m(\theta, \mathbf{s}^{m+1})] + {\frac{1-s_n^m}{2}} \log[1-p_n^m(\theta, \mathbf{s}^{m+1})])\label{13}
\end{equation}
If we plug in Equation (\ref{p}) and calculate its derivatives with respect to $\theta$ and $b$, we easily derive the \textit{local delta rule},
\begin{align}
    \frac{\partial F}{\partial \theta_{k,n}^{m+1,m}} &= -s_k^{m+1}(s_n^m-p_n^m) \\
     \frac{\partial F}{\partial b_n^{m+1,m}} &= -(s_n^m-p_n^m)
\end{align}
To update the recognition weights $\phi$ in the sleep phase, we exchange the relative positions of $P$ and $Q$, using $\tilde{F} = \mathbb{E}_{P_\theta}[\log P_\theta(\alpha,d)] - \mathbb{E}_{P_\theta}[\log Q_\phi(\alpha|d)]$ as a modified objective function, then the similar deductions follow which give the same local delta rules for recognition weights. Finally, all the parameters are updated by gradient descent $x = x - \gamma \frac{\partial f}{\partial x} $, where $\gamma$ is the learning step size. 

\begin{remark}
From the derivation of the local delta rule we can tell, by decoupling the forward and backward passes, and updating based on a single sampled instance, the objective function is simplified too much to be considered as variational or free-energy. However, this working algorithm is computationally cheap and efficient, and it approximates the true objective in an ensemble sense, which means the system minimizes the variational free energy after enough rounds of iterations with sufficient accuracy (probably converges to a local minima instead of the global minima).
\end{remark}

\begin{remark}
The local delta rule is the simplest updating rule we could derive from Equation (\ref{11}). If we keep the weighting term $Q_\phi(\alpha|d)$ in Equation (\ref{12}), the local delta rule will also be weighted by this term, which gives what we call the \textit{weighted local delta rule}. In \cite{dayan1995helmholtz}, another rule is given by replacing the stochastic neuron activities with their holistic mean, which is the calculated probabilities of neuron activations. In our preliminary experiments, we started with the classical local delta rules. The comparison and evaluation of all updating rules will be reserved for future work.
\end{remark}

Here we present a brief account of the two-phase training mechanism of the Helmholtz machine in comparison to the VAE. The Helmholtz machine is widely acknowledged as the predecessor of VAE and both neural networks resort to variational machine learning. Instead of alternative two-phase training, VAE uses a single objective function that jointly optimizes the recognition and generative processes. The advantage of the wake-sleep algorithm, for our application purposes, mainly lies in three areas:

\begin{enumerate}
\item The decoupling of recognition and generation spares a whole lot of space for creative manipulation and subtle mediation between these two processes, thus available the ground for studying computational creativity and foraging artistic usage of the model (future directions).
\item The idea of analysis-by-synthesis and inverting the hierarchical model in real-time make active inference and real-time modifications possible. One of the biggest differences between the Helmholtz machine and the VAE is that the generation in the Helmholtz machine is unconstrained. It's not subject to the maximum likelihood and any generated sample could be used to update the model parameters. In other words, the Helmholtz machine presents more flexibility in the training process, a desired feature to serve our purposes.
\item The hierarchical structure and forward-backward connections in the Helmholtz machine are a good parallel for the cortical hierarchies in our brain. This point be illustrated in more detail in the following arguments.
\end{enumerate}

\subsection{Hierarchical Model for the Brain}

It's contended extensively in FEP-related works that the cortical responses in our brain observe a hierarchical model with forward and backward connections, where the forward driving connections convey prediction errors from a lower area to a higher area, and nonlinear backward connections construct predictions \cite{friston2005theory} \cite{friston2008hierarchical} \cite{friston2009predictive}. The brain trying to infer the causes of its sensory inputs and generating corresponding sensations is also a prevailing idea in FEP. As the author doesn't have a real background in neurobiology, the statements in this subsection couldn't be presented with too much precision or detail. But in general, the idea of analysis-by-synthesis is well demonstrated in the Helmholtz machine, and Hinton pointed out that one of their motivations for developing the Helmholtz machine is inspired by Friston's cortical hierarchy theory \cite{hintonlecture}. In \cite{friston2009predictive}, Friston also cited the paper on the Helmholtz machine \cite{dayan1995helmholtz}. By this cross-referencing, we believe there is enough evidence to link the brain architecture to the Helmholtz machine, and study the brain functions such as predictive coding, message passing, and predictive processing via the Helmholtz machine, at least at a reasonable metaphorical level. Two preliminary comments could be made at this stage by the working mechanism of the Helmholtz machine, while the validation and more systematic discussions could only be realized after the numerical experiments are carried out and a detailed examination is applied in future work.

\begin{enumerate}
\item In predictive processing \cite{Clark2015RadicalPP}, the processing of sensory inputs goes both ways within the hierarchical model. Besides the bottom-up passive pattern recognition of the stimuli, the brain also actively constructs the predictions via top-down connections. The backward connections regulate the precision of prediction errors that allow the system to cope with noisy and ambiguous sensory inputs. In other words, the separate treatment of forward and backward connections, which correspond to the wake and sleep phases in the Helmholtz machine, is imperative to study PP (predictive processing) and PC (predictive coding) related brain functions. Besides, due to the functional asymmetry of forward and backward connections in the brain\cite{friston2009predictive}, where backward connections are more modulatory or nonlinear in their effects on neuronal responses, it's also advantageous to decouple the two processes and treat them separately, which allow more flexibility.

\item In hierarchical message passing, the synapses are characterized by their local efficacy. It means that the prediction errors are resolved by each neuron locally, which only receives responses from neurons at its current and preceding levels. This local efficacy well corresponds to the local delta updating rules derived in the Helmholtz machine. This condition is important because it permits a biologically plausible implementation, where the connections driving inference run only between neighboring levels \cite{friston2005theory}.

\end{enumerate}

\section{Experiment}

The preliminary experiment is designed by a 4-layer Helmholtz machine, with $10, 8, 5, 3$ neurons in each layer respectively (see Fig. \ref{fig2}). This section will present a detailed experimental setup with data design and two-stage training, as well as the experimental result which preserves adequate generative diversity, while boosting the generation accuracy above 0.99 in the meantime (please see the codes in my \href{https://github.com/Jovie-Liu/Helmholtz-machine-FEP}{GitHub}).

\subsection{Training Stage I: Experimental Setup}

\begin{figure}
\centering
\includegraphics[scale = 0.58]{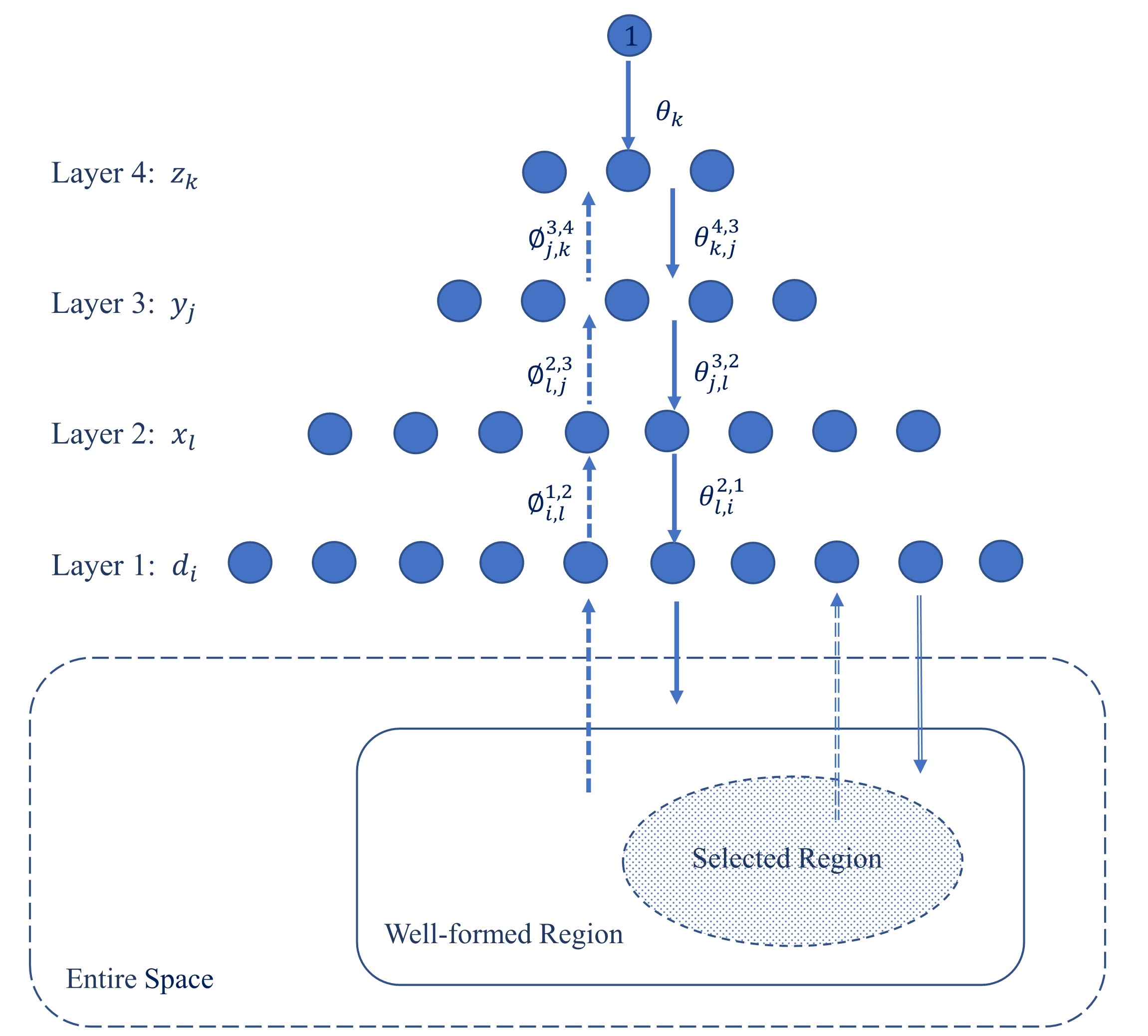}
\caption{Helmholtz Machine with Active Inference. The experiment is implemented with a 4-layer Helmholtz machine, with $10, 8, 5, 3$ neurons in each layer ascendingly. In the sleep phase, activities in layer 4 are generated by generative bias from unity. The training is implemented in two stages. In stage I, as the single lines connecting the training set and the model on the left side indicate, the inputs are from the well-formed region and the generations are unconstrained, which falls anywhere in the entire space; in stage II, the parameters are fine-tuned by restricting the generations within the well-formed region while actively deforming the input distribution based on the model generations (double lines on the right side), resulting in the selected region that conforms to the latent model representations.} \label{fig2}
\end{figure}

The data design is inspired by the phenotypic boundary discussed in \cite{friston2009reinforcement}. For a phenotype to exist it must possess defining characteristics or traits. These traits essentially limit the agent to a bounded region in the space of all states it could be in. Once outside these bounds, it ceases to possess that trait (cf, a fish out of water). This bounded region corresponds to the well-formed region in Fig. \ref{fig2} and the entire space represents all possible states of the world. 

To construct a valid subset of all $1024$ possibilities of the combination of binary-valued first-layer data neurons ($2^{10}$), we devise three well-formed rules inspired by the musical gestalt. Metaphorically, we consider the $10$ neurons as a $10$-note sequence, which represents a rhythmic pattern -- $0$ or $-1$ denotes the rest (we will use $0$ for discussion convenience but in numerical implementation it's always replaced by $-1$) and $1$ denotes a percussive attack. Then the rules entail, what is a valid rhythmic pattern?
\begin{description}
\item[Rule 1] The sequence always starts with $1$.
\item[Rule 2] Forbid single event that's strongly isolated from other groups ($00100$), and also avoid isolated event at the beginning ($100$) and the end ($001$) of the sequence.
\item[Rule 3] Forbid the extended break ($0000$).
\end{description}
We won't give further explanations for the designing logic for these rules, but we refer the readers who are interested to \cite{lerdahl1996generative} for a better understanding of musical grouping structures.

The advantage of rule-based generation is that, we have a metric to assess the goodness of the generated samples by simply checking them against the rules, thus the model performance could be measured with certainty. In training stage I, we use this generated well-formed set as inputs, and update the recognition weights with arbitrarily generated instances (see the single lines connecting the data and machine on the left side in Fig. \ref{fig2}). As explained in \cite{hinton1995wake}, the model aims to find the economical latent representations that prescribe the minimum description length. After sufficient iterations, the generation accuracy reached $0.94 \pm 0.01$, and couldn't be further improved by repeating the current training.

\subsection{Training Stage II: Active Inference}

In training stage II, we fine-tune the model trained in stage I by active inference, which boosted the generation accuracy to $0.99+$ with only $200$ rounds of iterations. In this stage, the generated instances in the sleep phase are filtered by the well-formed rules, thus only valid generations within the phenotypic bounds are accepted to train the recognition weights. In the meantime, the valid generations are maintained to actively modify the distribution of the input set. The more an instance is generated, the more salient it becomes in the evidence distribution. 

This distribution modification could be viewed either as salience\cite{parr2019attention}, in a similar way of executing eye movements to sample the sensations that conform to the agent's expectations; or as niche construction\cite{veissière_constant_ramstead_friston_kirmayer_2020}, that renders real modifications on the environment such as the "desire path". Either way, the data distribution changes due to active inference, thus the given input data (data in the well-formed region) becomes actively sampled data (data in the selected region) that reflects the current representations in the "brain" (see the double lines on the right side connecting the data and machine in Fig. \ref{fig2}).

\begin{figure}
\centering
\includegraphics[scale = 0.3]{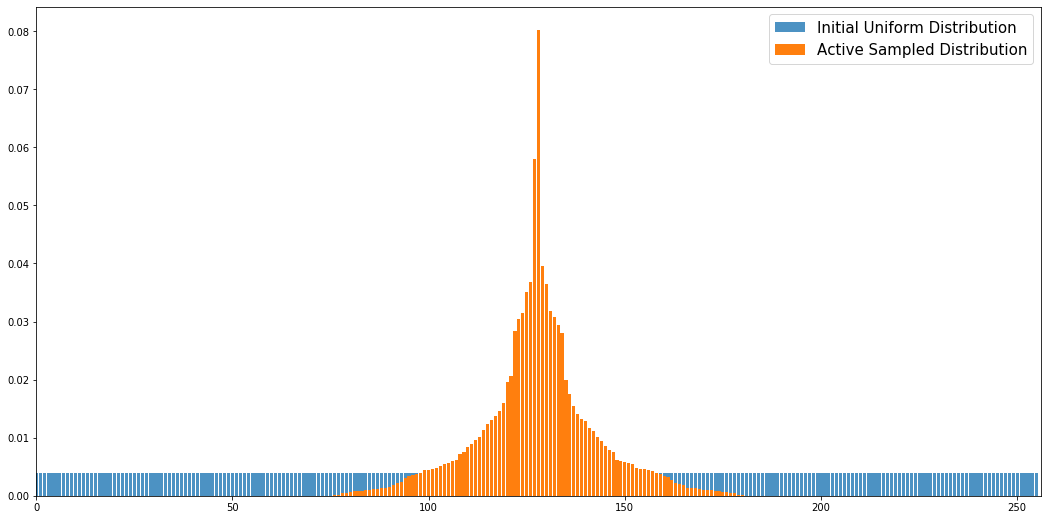}
\caption{Reconfigured Data Distribution by Active Sampling. The well-formed set gives a uniform initial distribution over all valid data points (in blue), which is continuously deformed to the distribution in orange by active sampling.} \label{fig3}
\end{figure}

The input data distribution is described in Fig. \ref{fig3}. The FEP-based active training continuously deforms the uniform equal-probability distribution of the initial dataset (in blue) to the active sampled distribution (in orange) that fits the internal representation and capacity of the machine. After this second-stage fine-tuning, the Helmholtz machine is able to generate almost 100\% accurate samples with a more focused range within all possibilities of the well-formed set while keeping generative diversity to a satisfactory degree.

\bibliographystyle{splncs04}
\bibliography{ref}

\begin{thebibliography}{10}
\providecommand{\url}[1]{\texttt{#1}}
\providecommand{\urlprefix}{URL }
\providecommand{\doi}[1]{https://doi.org/#1}

\bibitem{allen2018cognitivism}
Allen, M., Friston, K.J.: From cognitivism to autopoiesis: towards a
  computational framework for the embodied mind. Synthese  \textbf{195}(6),
  2459--2482 (2018). \doi{10.1007/s11229-016-1288-5}

\bibitem{2016arXiv160100670B}
{Blei}, D.M., {Kucukelbir}, A., {McAuliffe}, J.D.: {Variational Inference: A
  Review for Statisticians}. arXiv e-prints arXiv:1601.00670 (Jan 2016).
  \doi{10.48550/arXiv.1601.00670}

\bibitem{carhart2010default}
Carhart-Harris, R.L., Friston, K.J.: The default-mode, ego-functions and
  free-energy: a neurobiological account of freudian ideas. Brain
  \textbf{133}(4),  1265--1283 (2010). \doi{10.1093/brain/awq010}

\bibitem{Clark2015RadicalPP}
Clark, A.: Radical predictive processing. Southern Journal of Philosophy
  \textbf{53},  3--27 (2015). \doi{10.1111/sjp.12120}

\bibitem{constant2018variational}
Constant, A., Ramstead, M.J., Veissiere, S.P., Campbell, J.O., Friston, K.J.: A
  variational approach to niche construction. Journal of the Royal Society
  Interface  \textbf{15}(141),  20170685 (2018). \doi{10.1098/rsif.2017.0685}

\bibitem{dayan1995helmholtz}
Dayan, P., Hinton, G.E., Neal, R.M., Zemel, R.S.: The helmholtz machine. Neural
  computation  \textbf{7}(5),  889--904 (1995). \doi{10.1162/neco.1995.7.5.889}

\bibitem{demekas2020investigation}
Demekas, D., Parr, T., Friston, K.J.: An investigation of the free energy
  principle for emotion recognition. Frontiers in Computational Neuroscience
  \textbf{14}, ~30 (2020). \doi{10.3389/fncom.2020.00030}

\bibitem{friston2005theory}
Friston, K.: A theory of cortical responses. Philosophical transactions of the
  Royal Society B: Biological sciences  \textbf{360}(1456),  815--836 (2005).
  \doi{10.1098/rstb.2005.1622}

\bibitem{friston2008hierarchical}
Friston, K.: Hierarchical models in the brain. PLoS computational biology
  \textbf{4}(11),  e1000211 (2008). \doi{10.1371/journal.pcbi.1000211}

\bibitem{friston2010free}
Friston, K.: The free-energy principle: a unified brain theory? Nature reviews
  neuroscience  \textbf{11}(2),  127--138 (2010).
  \doi{https://doi.org/10.1038/nrn2787}

\bibitem{FRISTON2016862}
Friston, K., FitzGerald, T., Rigoli, F., Schwartenbeck, P., ODoherty, J.,
  Pezzulo, G.: Active inference and learning. Neuroscience and Biobehavioral
  Reviews  \textbf{68},  862--879 (2016).
  \doi{https://doi.org/10.1016/j.neubiorev.2016.06.022},
  \url{https://www.sciencedirect.com/science/article/pii/S0149763416301336}

\bibitem{friston2009predictive}
Friston, K., Kiebel, S.: Predictive coding under the free-energy principle.
  Philosophical transactions of the Royal Society B: Biological sciences
  \textbf{364}(1521),  1211--1221 (2009). \doi{10.1098/rstb.2008.0300}

\bibitem{doi:10.1080/17588928.2015.1020053}
Friston, K., Rigoli, F., Ognibene, D., Mathys, C., Fitzgerald, T., Pezzulo, G.:
  Active inference and epistemic value. Cognitive Neuroscience  \textbf{6}(4),
  187--214 (2015). \doi{10.1080/17588928.2015.1020053}, pMID: 25689102

\bibitem{friston2009reinforcement}
Friston, K.J., Daunizeau, J., Kiebel, S.J.: Reinforcement learning or active
  inference? PloS one  \textbf{4}(7),  e6421 (2009).
  \doi{10.1371/journal.pone.0006421}

\bibitem{friston2020generative}
Friston, K.J., Parr, T., Yufik, Y., Sajid, N., Price, C.J., Holmes, E.:
  Generative models, linguistic communication and active inference.
  Neuroscience \& Biobehavioral Reviews  \textbf{118},  42--64 (2020).
  \doi{10.1016/j.neubiorev.2020.07.005}

\bibitem{friston2018deep}
Friston, K.J., Rosch, R., Parr, T., Price, C., Bowman, H.: Deep temporal models
  and active inference. Neuroscience \& Biobehavioral Reviews  \textbf{90},
  486--501 (2018). \doi{10.1016/j.neubiorev.2017.04.009}

\bibitem{hintonlecture}
Hinton, G.E.: Lecture 13.4 - the wake sleep algorithm (2017),
  \url{https://www.youtube.com/watch?v=FBkhbqrFyo4&list=PLLssT5z_DsK_gyrQ_biidwvPYCRNGI3iv&index=63}

\bibitem{hinton1995wake}
Hinton, G.E., Dayan, P., Frey, B.J., Neal, R.M.: The" wake-sleep" algorithm for
  unsupervised neural networks. Science  \textbf{268}(5214),  1158--1161
  (1995). \doi{10.1126/science.7761831}

\bibitem{2013arXiv1312.6114K}
{Kingma}, D.P., {Welling}, M.: {Auto-Encoding Variational Bayes}. arXiv
  e-prints arXiv:1312.6114 (Dec 2013). \doi{10.48550/arXiv.1312.6114}

\bibitem{kirchhoff2018markov}
Kirchhoff, M., Parr, T., Palacios, E., Friston, K., Kiverstein, J.: The markov
  blankets of life: autonomy, active inference and the free energy principle.
  Journal of The royal society interface  \textbf{15}(138),  20170792 (2018).
  \doi{10.1098/rsif.2017.0792}

\bibitem{koelsch2019predictive}
Koelsch, S., Vuust, P., Friston, K.: Predictive processes and the peculiar case
  of music. Trends in cognitive sciences  \textbf{23}(1),  63--77 (2019).
  \doi{10.1016/j.tics.2018.10.006}

\bibitem{lerdahl1996generative}
Lerdahl, F., Jackendoff, R.S.: A Generative Theory of Tonal Music, reissue,
  with a new preface. MIT press (1996)

\bibitem{mazzaglia2022free}
Mazzaglia, P., Verbelen, T., {\c{C}}atal, O., Dhoedt, B.: The free energy
  principle for perception and action: A deep learning perspective. Entropy
  \textbf{24}(2), ~301 (2022). \doi{10.3390/e24020301}

\bibitem{parr2019attention}
Parr, T., Friston, K.J.: Attention or salience? Current opinion in psychology
  \textbf{29}, ~1--5 (2019). \doi{10.1016/j.copsyc.2018.10.006}

\bibitem{parr2019generalised}
Parr, T., Friston, K.J.: Generalised free energy and active inference.
  Biological cybernetics  \textbf{113}(5-6),  495--513 (2019).
  \doi{10.1007/s00422-019-00805-w}

\bibitem{tschantz2020reinforcement}
Tschantz, A., Millidge, B., Seth, A.K., Buckley, C.L.: Reinforcement learning
  through active inference. arXiv preprint arXiv:2002.12636  (2020).
  \doi{10.48550/arXiv.2002.12636}

\bibitem{veissière_constant_ramstead_friston_kirmayer_2020}
Veissière, S.P.L., Constant, A., Ramstead, M.J.D., Friston, K.J., Kirmayer,
  L.J.: Thinking through other minds: A variational approach to cognition and
  culture. Behavioral and Brain Sciences  \textbf{43}, ~e90 (2020).
  \doi{10.1017/S0140525X19001213}

\end{thebibliography}
%




\end{document}